\pdfoutput=1
\documentclass[letterpaper, 10 pt, conference]{ieeeconf}  

\IEEEoverridecommandlockouts                              

\usepackage{times}
\usepackage{epsfig}
\usepackage{graphicx}
\usepackage{amsmath}
\usepackage{amssymb}

\usepackage{graphicx}
\usepackage{amsmath}
\usepackage{amssymb}
\usepackage{booktabs}
\usepackage[dvipsnames]{xcolor}
\usepackage{multicol}
\usepackage{multirow}
\usepackage{xcolor}
\usepackage{float}
\usepackage{dsfont}
\usepackage{subcaption}

\title{\LARGE \bf
Scaling Motion Forecasting Models with Ensemble Distillation
}

\author{Scott Ettinger$^{1*}$ \and Kratarth Goel$^{1*}$ \and  Avikalp Srivastava$^{1}$  \and Rami Al-Rfou$^{1}$
\thanks{*Indicates equal contributions}
\thanks{$^{1}$Research scientist at Waymo; contact: settinger@waymo.com}%
}

\begin{document}

\maketitle
\thispagestyle{empty}
\pagestyle{empty}

\begin{abstract}
Motion forecasting has become an increasingly critical component of autonomous robotic systems. Onboard compute budgets typically limit the accuracy of real-time systems.
In this work we propose methods of improving motion forecasting systems subject to limited compute budgets by combining model ensemble and distillation techniques.
The use of ensembles of deep neural networks has been shown to improve generalization accuracy in many application domains.
We first demonstrate significant performance gains by creating a large ensemble of optimized single models.
We then develop a generalized framework to distill motion forecasting model ensembles into small student models which retain high performance with a fraction of the computing cost. For this study we focus on the task of motion forecasting using real world data from autonomous driving systems.
We develop ensemble models that are very competitive on the Waymo Open Motion Dataset (WOMD) and Argoverse leaderboards. From these ensembles, we train distilled student models which have high performance at a fraction of the compute costs. These experiments demonstrate distillation from ensembles as an effective method for improving accuracy of predictive models for robotic systems with limited compute budgets.
\end{abstract}

 \section{Introduction}

Autonomous robotic systems require high quality motion forecasting of dynamic objects in the scene in order to produce safe and optimal planning routes. The accuracy of forecasting systems is limited by the computational budget of the hardware running onboard the robot. In this work, we propose a method using ensemble distillation to scale performance beyond this limit and improve accuracy within the same compute budget.

In the seminal 2015 paper, Hinton et al. \cite{Hinton2015DistillingTK} proposed the method of model distillation, which is a process of transferring knowledge from a large complex model into a simpler one.
Knowledge distillation has become a de facto standard to improve the performance of small neural networks.
By first training a more expensive high accuracy ``teacher" model, the accuracy of a smaller ``student" model can be improved by training the same model but using information from the teacher model outputs through additional distillation losses. In this context, we propose to use distillation to train student models suitable for running onboard autonomous systems in order to significantly improve quality.

\begin{figure}[h]
    \centering
    \includegraphics[width=0.8\linewidth]{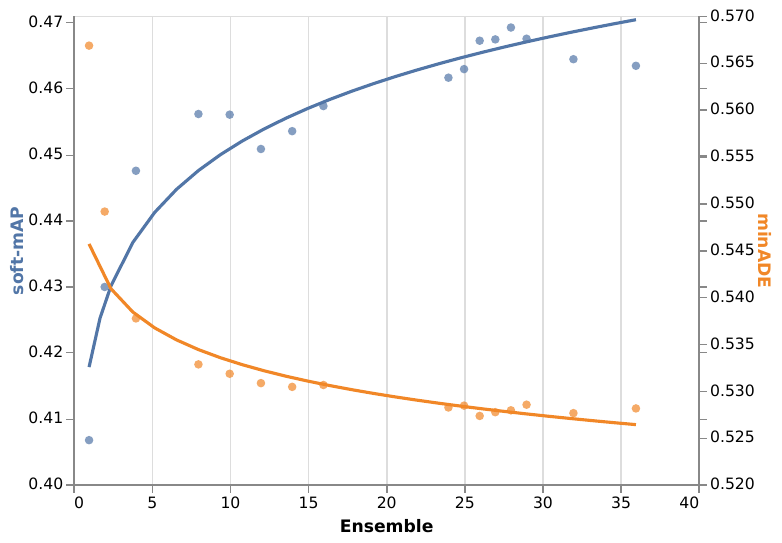}
    \caption{ Ensembles provide a method to improve both the soft-mAP and the minADE metrics for trajectory prediction with a linear increase in compute. But can we do better and achieve similar scaling of performance without the additional compute cost? }
    \label{fig:ensemble_metrics}
\end{figure}

To provide a high accuracy teacher model for distillation we propose using an ensemble of multiple independently trained motion forecasting models. Using ensemble models is a well established technique to improve the performance of deep learning models. It has been shown that ensembling the outputs of identical models trained on the same data can provide significant performance improvements\cite{Allen-Zhu2020Distill}. Individual component models of an ensemble can be trained in parallel to create a model with much larger capacity without increasing training time. Ensembles avoid over-fitting by providing better bias-variance trade-offs as compared to training a single large model with more parameters.
Motion forecasting models must output a distribution of possible trajectories for each agent as there are multiple valid futures for each prediction.
Typically the modes of the distribution outputs do not have any correspondence across different models in an ensemble so simple averaging techniques cannot be used to create an ensemble.
We propose a method for combining any set of heterogeneous motion forecasting models into a higher performing ensemble. Results of this method are shown in Fig \ref{fig:ensemble_metrics}, where both minADE and soft-mAP performance metrics improve with more teacher models in the ensemble.

While ensembles can provide significantly higher accuracy, the high compute cost makes it impractical to deploy them in an onboard autonomous system. We propose a method of using distillation to bring the cost back down to within the onboard compute budget while retaining high performance. While distillation has been studied in many domains, distillation of motion forecasting models has its own unique challenges. Since motion forecasting models output not just a classification, but a distribution of trajectories, the distillation process must be able to train the student model to match the teacher model distribution. Another challenge is to balance the diversity of outputs of the distilled model to handle longer tail cases while retaining good performance on common cases. We develop a general framework to distill an ensemble that can be applied to a heterogeneous set of teachers. Our main contributions can be summarized as follows:
1. We develop a general framework for ensemble distillation of motion forecasting models.
2. We achieve very competitive results on the Waymo Open Dataset (WOMD) and Argoverse 2 marginal prediction challenges with an ensemble model.
3. We train a distilled model which provides high performance at a significantly reduced compute cost.

\section{Related Work}

\paragraph{Motion Forecasting}
The recent availability of large scale autonomous driving datasets and benchmarks  \cite{argoverse,lyft5,WOMD} have helped drive interest in motion prediction tasks. 
One class of models draws inspiration from the computer vision literature, rendering inputs as a multichannel rasterized top-down image \cite{cui2019mtp,sapp2019multipath,DBLP:journals/corr/abs-1906-08945,DBLP:journals/corr/abs-2101-07907,trajectronpp,zhao2020tnt}.
A natural language inspired popular alternative is to model the scene entities as tokens of a language.
With this approach, agent state history is typically encoded via sequence modeling techniques like RNNs \cite{multiheadAF,wimp2020,social-lstm,rhinehart2019precog} or temporal convolutions \cite{liang2020laneGCN}.
Road elements are approximated with basic primitives (e.g. piecewise-linear segments) which encode pose information and semantic type.
Modeling relationships between entities is often presented as an information aggregation process, and models employ pooling \cite{zhao2020tnt,gao2020vectornet,social-lstm,social-gan,multiheadAF,DBLP:journals/corr/LeeCVCTC17}, soft-attention \cite{multiheadAF,zhao2020tnt} as well as graph neural networks \cite{casas2020spagnn,liang2020laneGCN,wimp2020}
Wayformer \cite{Nayakanti2022Wayformer} builds on top of transformers \cite{vaswani2017attention},  to increase the performance even further. This architecture serves as the baseline for experiments in this paper. 

\paragraph{Ensembling}
The scaling laws in \cite{scalinglaws1}  detail that it is often a combination of scaling up model size and compute as well as data that results in performance improvements. If data or compute is limited, scaling just the model size results in inefficiency and can also result in sub optimal results due to high variance and over fitting. Ensembling refers to the idea of constructing a single model from a set of weaker models \cite{Hansen1990NeuralNE,Dietterich2000EnsembleMI,Krogh1994NeuralNE}.
To learn a diverse set of models, several strategies of sampling the training dataset have been proposed such as bagging and boosting \cite{Maclin1999PopularEM,Breiman2004StackedR,Schapire1997BoostingTM}.
The resulting model not only has higher accuracy, but also more robust to out of distribution examples, and better at estimating uncertainty.
Deep learning ensembles typically rely on having random initialization seeds to learn a diverse set of models \cite{Fort2019DeepEA}.
Traditionally there are three different flavors of ensembling to scale up the model performance:

Model ensemble: True model ensembling consists of training $N$ separate copies of the model initialized with random weights to encourage diversity.  Each model can have an identical architecture and loss function or have specificity which makes them better at different things within a task. For simplicity we use this method with uniform models and loss functions in this work.

Bootstrap ensembling of the predictor heads: In this technique we specifically apply bootstrap aggregation (bagging) \cite{Maclin1999PopularEM} to our predictor heads by training $E$ such heads together in the same model. The heads can either be uniform as seen in \cite{Nayakanti2022Wayformer} or have varied architectures and loss functions \cite{shi2022mtr}. To encourage models to learn complementary information, the weights of the $E$ heads are initialized randomly.

Hybrid Ensemble: Here we can combine the above two approaches such that each model has $E$ predictor heads and we train $N$ replicas of them.

\paragraph{Knowledge Distillation}
Distillation is a technique where a smaller student model is trained using the outputs or intermediate representations of a larger teacher model.
While the mechanism is complex \cite{allen-zhu2020towards}, experimental results from many applications show that the performance of a larger (teacher) model can be captured by a smaller (student) model \cite{Hinton2015DistillingTK,Li2014LearningSD}.
It has also been shown that "self-distillation" from the same model can outperform the original model trained on ground truth data alone \cite{Furlanello2018BornAN}.
This hints to either improved generalization of the second generation teacher models or simpler optimization regime \cite{Zhou2020UnderstandingKD}. 

\paragraph{Structure Prediction Distillation}
While the original distillation work was applied to classification, follow up work proposed extending the technique for sequence generative models.
Applied to machine translation, \cite{Kim2016SequenceLevelKD} showed that improvements in the teacher proposed translations will carry over to the trained student models.
Recently, distillation has been applied to motion forecasting to improve the quality of a scene-centric student model using predictions from an ego-centric teacher \cite{Su2022NarrowingTC}. Motion prediction poses a challenge to the distillation problem, as each model outputs a complete distribution of possible futures. In this paper we develop a framework for distillation of the complete output distribution from teacher models to students in contrast to earlier approaches.

\paragraph{Ensemble Distillation}
While single teacher distillation to a student is well studied, distilling an entire ensemble of teacher models has only gained interest recently. \cite{allen-zhu2020towards} studies distillation of ensembles created with simple averages of independently trained identical models for classification. 
They provide theoretical results for ensemble distillation accuracy.
\cite{Malinin2019Ensemble} develops a method of ensemble distillation which preserves the ability to estimate uncertainty as applied to image classification networks. In this work we develop a technique to ensemble output distributions from different teacher models that have no correspondence between outputs and to distill the resulting ensembles into smaller student models using a novel distillation loss formulation.

\section{Modeling}
\label{sec:modeling}

In this work, we focus on the general modeling task of motion forecasting for agents in real world driving environments. Agents include vehicles, pedestrians, and cyclists.  The modeling task is to predict the future trajectory of each agent for a fixed time horizon given a fixed input window of agent histories in the scene.  

\begin{figure}[ht]
    \centering
    \includegraphics[trim={0 2cm 2cm 1cm},width=0.9\linewidth]{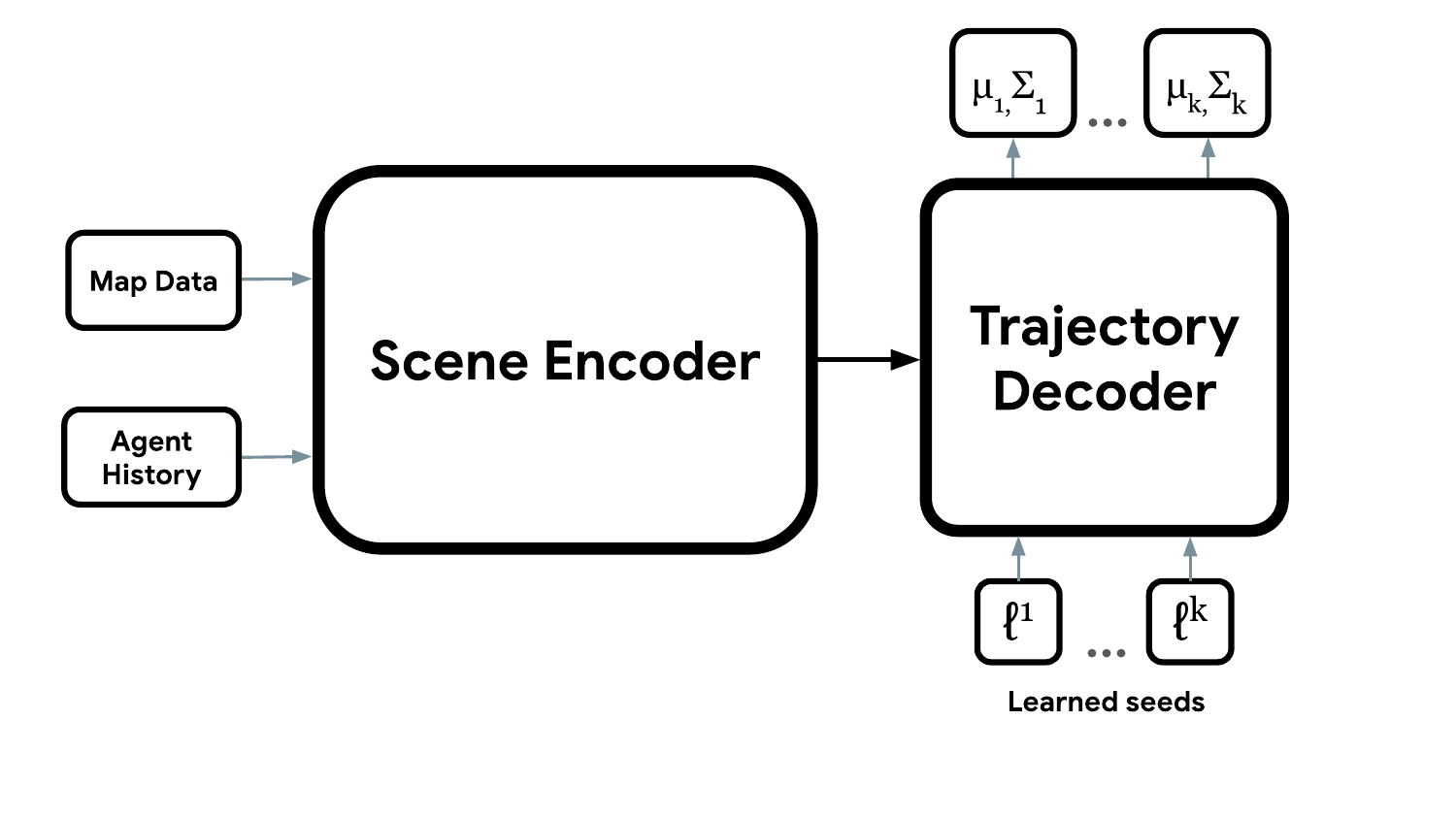}
    \caption{The Wayformer architecture is a pair of encoder/decoder Transformer networks. This model takes multimodal scene data as input and produces a multimodal distribution of trajectories.}
    \label{fig:wayformer}
\end{figure}

The nature of motion forecasting in this setting is inherently multi-modal as human agents in the scene can choose a variety of trajectories depending on their unobserved internal state. To account for this, the models output a distribution of trajectories for each agent.

The base motion forecasting model used in the experiments follows the Wayformer model described in \cite{Nayakanti2022Wayformer} as described in Figure \ref{fig:wayformer}. This model employs an attention based scene encoder and decoder with early fusion of the input modalities (map data, traffic light states, and agent motion boxes). The model outputs a Gaussian mixture model consisting of $K$ gaussian trajectory models each consisting of 2 dimensional mean positions $(x, y)$ and standard deviations for each time step.

The input to the model is comprised of multiple modalities that encode both the scene information as road lanes and traffic lights as well the dynamic agent information including agent state history and interactions.



The baseline model was tuned and optimized to maximize the performance achieved by a single model without over-fitting. A set of multiple instances of this model were independently trained to be used as the basis for the ensemble models described below.

\section{Ensemble Distillation}
\subsection{ Ensemble models }
\begin{figure}[h]
  \centering
   \includegraphics[trim={8cm 12cm 4cm 0cm},width=0.8\linewidth]{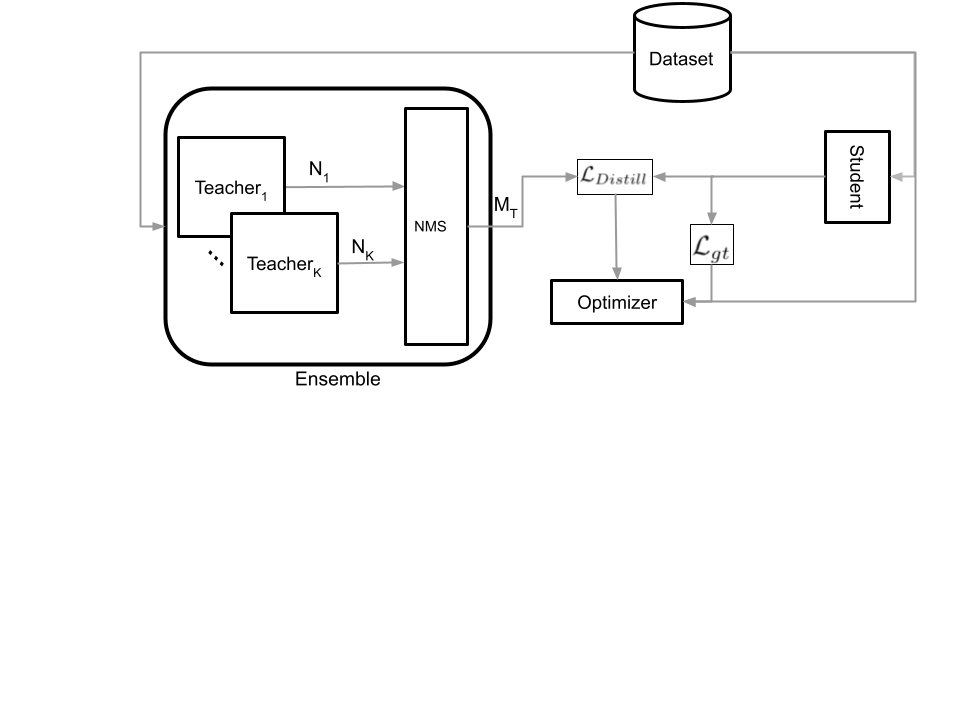}
   \caption{Illustration of the ensemble distillation pipeline. A set of $K$ teachers and a NMS that outputs $M_T$ trajectories form the ensemble. The student is trained with a groundtruth loss ($\mathcal{L}_{gt}$) and a distillation loss ($\mathcal{L}_{Distill}$). }
   \label{fig:pipeline}
\end{figure}

To create ensembles, model outputs are often combined together with simple averaging. However, due to the multi-modal nature of motion forecasting models, output modes from independently trained models will not correspond since they are learned independently during training.
To address this, We develop a general framework for ensemble distillation of motion forecasting models described in Fig \ref{fig:pipeline}. Notably, this framework can be applied to create and distill ensembles from models with different numbers of trajectory outputs and different architectures.

The output of each individual model within the ensemble is a Gaussian mixture model (GMM) for each agent as described in \cite{sapp2019multipath}.
Let $s_t$ denote the 2 dimensional position of an agent at time $t$, and $\mathbf{s}=[s_1, ..., s_T]$ is a trajectory to a fixed time horizon $T$.
Each individual model outputs a vector of parameters ${\boldsymbol{\Phi}}$ consisting of ${\boldsymbol{\pi}}$, $\boldsymbol{\mu}$, and $\boldsymbol{\Sigma}$ vectors to define a GMM model. The probability of a trajectory is given by
\begin{align} \label{eq:trajprob}
    p(\bold{s}\mid \boldsymbol{\Phi}) =&  \sum_{n=1}^N \pi^n \prod_{t=1}^T \mathcal{N}(s_t \mid \mu_t^n, w_{var} \Sigma_t^n)
\end{align}
where $N$ is the total number of modes/trajectories output by the model, $\mu_t^n$ and $\Sigma_t^n$ define a 2 dimensional multivariate Gaussian at each time step for $n$-th trajectory.  The mixing coefficients ${\pi^n}$ are fixed across all times $T$ and sum to one. 
\begin{align}
\sum_{n=1}^N \pi^n =& 1,  \pi^n \geq 0 \;\forall n
\end{align}
We introduce a scale factor $w_{var}$ which is applied to the variance of the model. This is typically set to less than one. If we set $w_{var}$ to $0$, we can sample the distribution by selecting one of the means $\mu_n$ based on the probabilities $\pi$.
Following \cite{Hinton2015DistillingTK}, a temperature adjustment $\tau$ is first applied to the mixture coefficients of each teacher model in order to boost the knowledge transfer of lower probability trajectories. These are computed per teacher as:
\begin{align}
    \pi^n =& \frac{e^{(log(\pi^n) / \tau)}}{\sum_{n=1}^N e^{(log(\pi^n) / \tau)}}
\end{align}

An ensemble distribution of $K$ individual teacher models can be created as a weighted sum of the probabilities of each model for a given sample $\bold{s}$:
\begin{align}
q(\bold{s} \mid \boldsymbol{\Phi_{ens}}) =& \sum_{k=1}^K w_k p(\bold{s}\mid \boldsymbol{\Phi_k})
\end{align}
where the weights $w_k$ sum to 1. Note that the number of modes $N$ does not need to be the same for all of the individual models in the ensemble.

\subsection {Distillation losses}
\label{section:losses}
In order for the distillation student model to mimic the teacher model, we utilize the cross entropy loss. To minimize the loss, the student model should maximize the probability of samples drawn from the teacher ensemble distribution $q(\bold{s} \mid \boldsymbol{\Phi_{ens}})$.
To train the student model, a set of samples are drawn from the teacher ensemble distribution and the student distillation loss term is computed as the negative log likelihood of the product of the probabilities of the samples:
\begin{align} \label{eq:sampling}
\bold{\hat{s}}_j \sim q(\bold{s} \mid \boldsymbol{\Phi_{ens}})
\end{align}
\begin{align} \label{eq:distillNLL}
\mathcal{L}_{DistillNLL} =& -log\prod_{j} p(\bold{\hat{s}_j} \mid \boldsymbol{\Phi_{Student}})
\end{align}
Minimizing this loss by varying $\boldsymbol{\Phi_{Student}}$ will maximize the student probability of the set of teacher samples. By doing so, the student model outputs become similar to that of the teacher.

The final distillation loss is computed as: 
\begin{align}
\mathcal{L}_{Total} =& \mathcal{L}_{DistillNLL} +  w_{gt}  \mathcal{L}_{gt}
\end{align}
where $\mathcal{L}_{gt}$ is the groundtruth loss used when training the model without distillation. We follow the same formulation as \cite{Nayakanti2022Wayformer} for the groundtruth loss. 

\subsection{ Non-Maximal Suppression }
\label{subsec:nms}
 The distillation pipeline is described in Fig \ref{fig:pipeline}. For a large ensemble of $K$ models each outputting $N_T = \sum^K_i N_i$ trajectories, 
 it may be too costly to store these trajectories or compute the loss over all of them during training.
 To reduce this, we can combine the teacher distributions into a single, smaller distribution prior to distillation training.
 Due to the multi-modal nature of the models, a simple averaging technique cannot be used as the outputs do not correspond directly across different models.
 In order to aggregate the distributions across all ensemble models we apply the method of non-maximal suppression (NMS) as described in \cite{multipathpp}.
 This method produces a fixed set of modes from the ensemble distributions while still preserving the diversity of the original output mixture.

The NMS algorithm first runs a clustering algorithm which greedily selects the output mode that covers the maximum total likelihood, and proceeds until all output modes have been covered.
The algorithm then runs an iterative refinement stage similar to KMeans clustering starting from the initial set of modes.
This produces a set of distribution parameters for a single mixture model with a reduced number of trajectories that retains the diversity of the original ensemble. Similarly, NMS can be applied to reduce the number of outputs from a single model.

\subsection {Efficient sampling of teacher models}
\label{section:approximation}

Evaluating a large number of samples from the teacher model for distillation may be too costly, even after reducing the teacher model with NMS.
An approximation can be performed by reducing the uncertainty of the teacher models by setting $w_{var}$ to zero in equation $\ref{eq:trajprob}$.
In this case, evaluating samples from the teacher reduces to using the mean values of each trajectory weighted by their probabilities. We use this simplification for all of our experiments.

\subsection{Mapping between teacher and student modes}  \label{section:special}
Another level of simplification to the loss function can be made when the number of modes output by the student $N_S$ and by the teacher $N_T$ are equal. In this case, we can simplify training by using a bijection from the modes of the teacher to those of the student and computing only corresponding losses. We apply a cross entropy loss between the teacher and student mixing coefficients to train the student probabilities. We denote the resulting student coefficients as $\pi_{1:1}$.
\section{Experimental Setup}
\label{sec:experiments}

\subsection{ Datasets \& Metrics}
For experiments we use two datasets: (1) the Waymo Open Motion dataset (WOMD) \cite{Ettinger2021WOMD} which contains over 570 hours of driving data including maps, traffic light states, and agent motion data consisting of 3D bounding boxes sampled at 10 Hz; (2) the Argoverse 2 Motion Forecasting dataset \cite{wilson2023argoverse} which consists of 250,000 scenarios mined for challenging interactions between the autonomous vehicle and other actors in each local scene, including maps and agent motion data sampled at 10Hz.  The WOMD task is to predict an 8 second future trajectory for up to 8 agents given a 1 second history of agent trajectories, whereas the Argoverse task is to predict a 6 second future trajectory for a single `focal' agent per scenario given 5 seconds of past history. For WOMD training, examples with classified u-turn and right and left turn trajectories were duplicated at a rate of 5\% to improve performance on rarer trajectories. 

\label{sec:metrics}
All of our evaluated models output 6 trajectories and all metrics are computed with $k=6$. We compare models using the following challenge metrics associated with the WOMD \cite{Ettinger2021WOMD} and AV2MF \cite{wilson2023argoverse} tasks, namely -  ${MR^t}$, ${minADE_k}$, ${mAP}$, ${soft-mAP}$, ${Overlap^t}$, ${minFDE}$, and ${Brier-minFDE}$.


\subsection{ Training Details and Hyperparameters }
For our experiments we chose the Wayformer model \cite{Nayakanti2022Wayformer} as the base architecture. The architecture follows the early-fusion wayformer variant as described in \cite{Nayakanti2022Wayformer}. We set the hidden size for all transformer layers to be 256. We use 2 encoder layers and 8 decoder layers. 


For all teacher and student model experiments, we train models using the AdamW optimizer \cite{adamw} with an initial learning rate
of 2e-4 with linear decay to 0 over 1M steps. We train models using 16 TPU v3 cores each, 
a batch size of 16 per core, resulting in a total batch size of 256 examples per step. All teacher models produce GMM distributions with $N_T = 64$ trajectories per agent. All settings are shared between teacher and student models except for the number of trajectory outputs. To create an N-model teacher ensemble, We train $K$ replicas of the teacher model described above with random initialization. The trajectories of all the teacher models are aggregated with non maximal suppression (section \ref{subsec:nms}) to generate a dataset with $M_T$ trajectories per agent output for each example in the input dataset. For distillation we then use the distillation loss defined in (section \ref{section:losses}). The students output $N_S$ number of trajectories which are optionally passed through an NMS pipeline, same as the teachers that aggregates these to $M_S$ trajectories.

\begin{figure*}[h]
\centering
    \includegraphics[width=\textwidth]{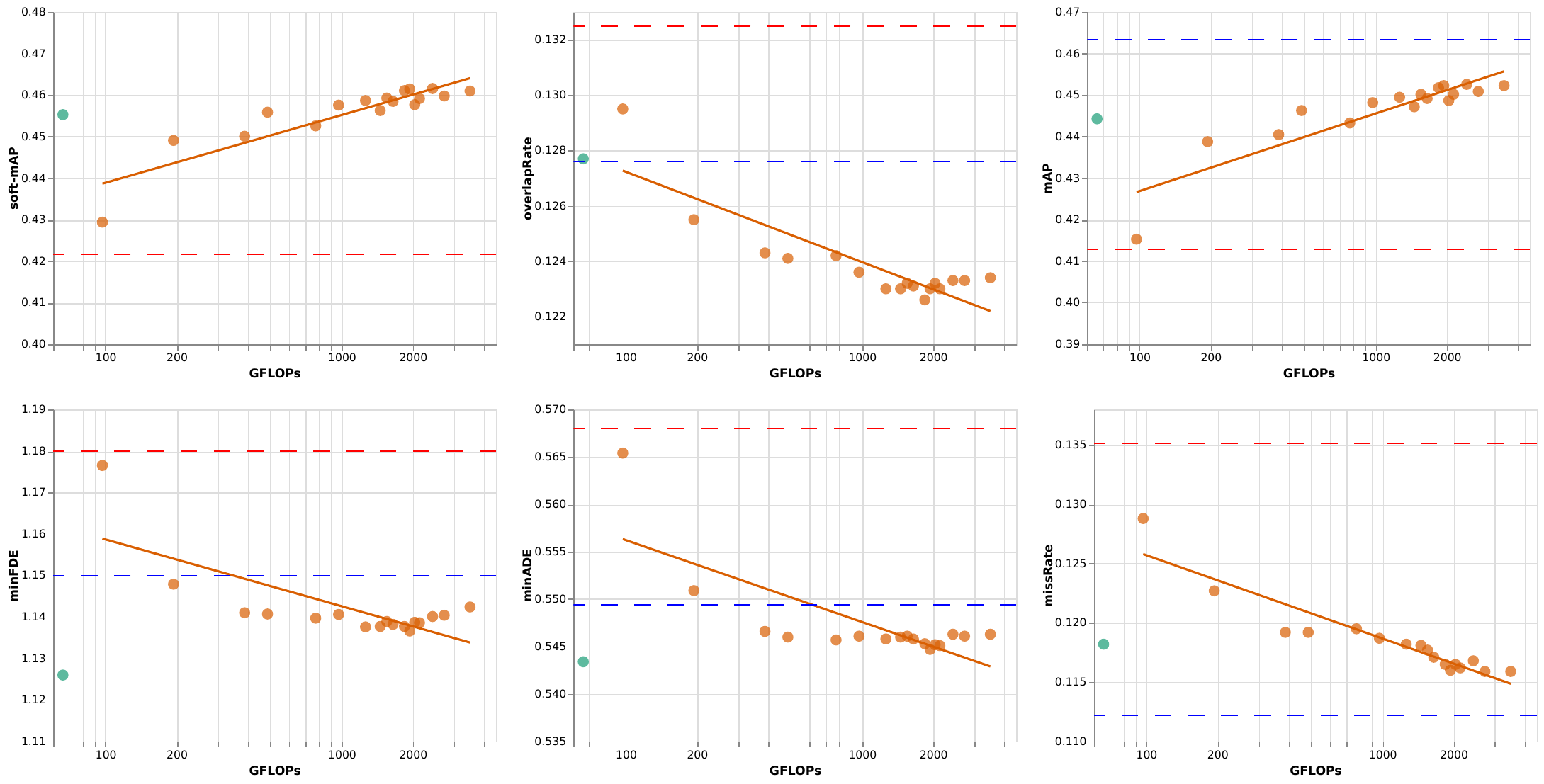}
\caption{We compare the metrics for Waymo Open Motion Dataset for our ensemble models and the distilled student models as a function of inference FLOPs (total floating point operations) relative to the single Wayformer \cite{Nayakanti2022Wayformer} single head model.  The relative FLOPs on the x-axis are shown in log scale. The Ensemble models are represented in \textbf{\textcolor[HTML]{F26035}{orange}}, with ensemble size $K$ linearly related to the FLOPs. Distilled student models are represented in \textbf{\textcolor[HTML]{009B55}{green}}. The \textbf{\textcolor{blue}{blue}} line represents the  ensembling SoTA results and non-ensemble SoTA is represented by the \textbf{\textcolor{red}{red}} line. } \label{fig:4pics}
\end{figure*}

\begin{table*}[h]
    \centering
    \resizebox{\textwidth}{!}{


\begin{tabular}{l|cccccc|c}
\toprule
 & \multicolumn{5}{c}{\textbf{Waymo Open Motion Dataset}}  \\ 
Models & minFDE ($\downarrow$) & minADE  ($\downarrow$) & MR  ($\downarrow$)  & Overlap ($\downarrow$) & mAP ($\uparrow$)  & soft-mAP$^{*}$ ($\uparrow$) & rel-FLOPs($\downarrow$) \\ 
\midrule

DenseTNT \cite{gu21dense_tnt}  & 1.551 & 1.039 & 0.157 & 0.178   & 0.328 & - & -\\
MultiPath \cite{multipath}  & 2.040 & 0.880 & 0.345 & 0.166 & 0.409 &  - & -\\
MultiPath++ \cite{multipathpp}  & 1.158  & 0.556  & 0.134  & 0.131 & 0.409 & -  & -\\
Wayformer \cite{Nayakanti2022Wayformer} & 1.126 & 0.545  & 0.123 & 0.127 & 0.412 & - & 1x \\
Wayformer Multi-Head \cite{Nayakanti2022Wayformer}  & 1.128 & 0.545 & 0.122 & 0.127 & 0.419 & 0.433 & 1.96x  \\
GTR-R36 & 1.223 & 0.601 & 0.133 & 0.128 & 0.426 & 0.438 & -\\
MTR++ & 1.194 & 0.591 & 0.130 & 0.128 & 0.433 & 0.441 & -\\
IAIR+ & 1.168 & 0.578 & 0.124 & 0.126 & 0.435 & 0.448 & -\\
Wayformer Student (Ours)   & \textbf{1.122} & \textbf{0.546} & \textbf{0.117} & \textbf{0.127} & \textbf{0.438} & \textbf{0.449} & \textbf{1.36x} \\
EDA\_single & 1.170 & 0.572 & 0.117 & 0.127 & 0.440 & 0.451 & -\\
GTR\_ens & 1.206 & 0.586 & 0.130 & 0.128 & 0.443 & 0.452 & -\\
Wayformer Ensemble 20 (Ours)  & \textbf{1.137} & \textbf{0.549} & \textbf{0.118} & \textbf{0.124} & \textbf{0.446} & \textbf{0.455} & \textbf{20x} \\
MGTR & 1.214 & 0.592 & 0.130 & 0.128 & 0.451 & 0.460 & -\\
MTR++\_Ens & 1.117 & 0.558 & 0.112 & 0.128 & 0.463 & 0.474 & -\\
\bottomrule
\end{tabular}

    }
    \caption{Comparison of results with the Waymo Open Motion Dataset 2022 leaderboard. \textbf{*} denotes the metric used for leaderboard ranking. rel-FLOPs indicates inference compute requirement relative to the Wayformer \cite{Nayakanti2022Wayformer} single head model.}
    \label{tab:sota}
\end{table*}

\section{Results}

We first study the effect of ensembles and distillation on the Waymo Open Motion Dataset \cite{WOMD} validation set as shown in Figure \ref{fig:4pics}. The best performing ensemble and distilled model are evaluated on the WOMD test set leaderboard and compared against other competitive methods in Table \ref{tab:sota}. All metric computations use the means of the model trajectory distribution outputs as predictions.

\subsection{Ensembles}
The orange dots and lines in figure \ref{fig:4pics} show the WOMD validation metrics and trends of results for ensemble models as a function of the FLOPs (total floating point operations) required to run inference on the model. The charts show that the performance follows a positive log-linear scaling relationship with FLOPs for all metrics on the WOMD leaderboard.

\begin{figure}[h]
\centering
\label{fig:argoverse}
    \includegraphics[width=\columnwidth]{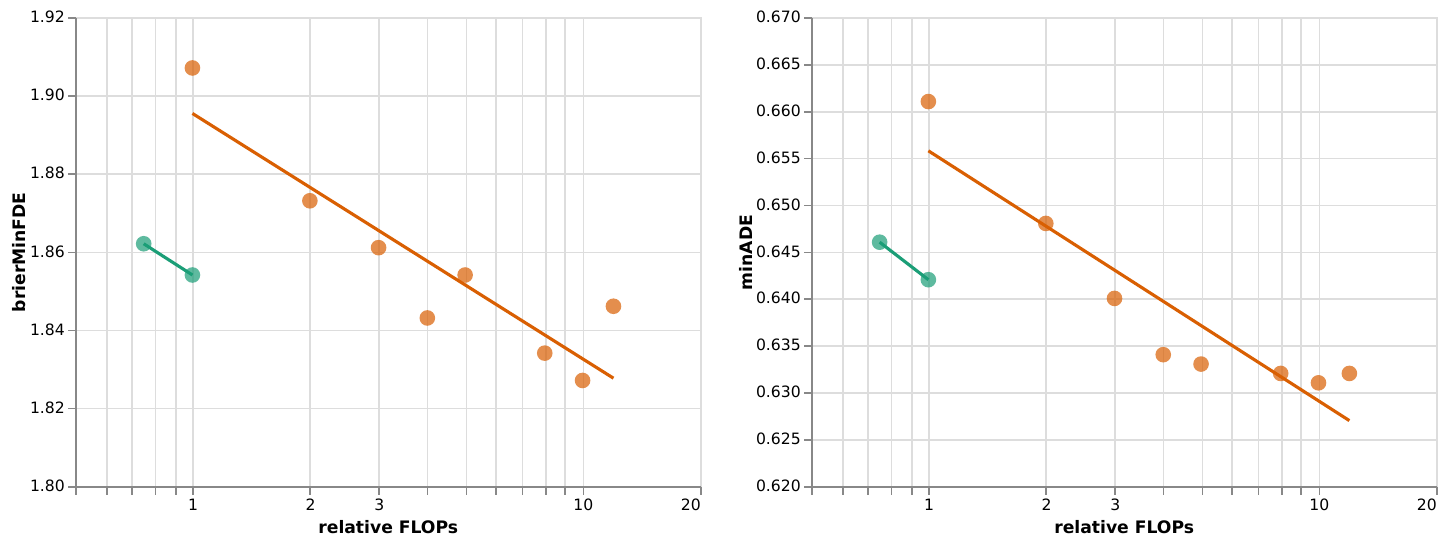}
\caption{Scaling ensemble result using the Argoverse dataset. The Ensemble models are represented in \textbf{\textcolor[HTML]{F26035}{orange}}, with ensemble size $K$ linearly related to the FLOPs. Distilled student models are represented in \textbf{\textcolor[HTML]{009B55}{green}}.} \label{fig:argoverse}
\end{figure}

It performs particularly well on minADE and minFDE metrics relative to the state of the art. Our ensemble of $K = 20$ teacher models achieves competitive performance for all metrics, placing third on the WOMD leaderboard. Future experiments will update this figure with distilled models with larger capacity.
Experiments on the Argoverse dataset shown in Fig \ref{fig:argoverse} show a similar positive log-linear scaling relationship with FLOPs for ensembles where the orange dots and lines represent the results for the ensemble models. The 10 model ensemble places 3rd on the Argoverse leaderboard. These results importantly, validate the hypothesis that ensembling can provide a way to scale model performance on motion forecasting problems. 
We also analyze qualitative improvements to model outputs in Fig \ref{fig:qualitative}, with examples of prediction quality of the ensemble model.

\subsection{Ensemble Distillation} \label{subsec:distill}
The green dots and lines in figure \ref{fig:4pics} show the WOMD validation metrics and trends of results for distillation models as a function of FLOPs. All distillation models are distilled from the $K=20$ model ensemble as described above using the efficient sampling procedure described in \ref{section:approximation} and setting $\tau=8$, $w_{gt}=0.4$, and $w_{var}=0.5$. The first result to note is that the distilled model (the far left green point) outperforms its corresponding baseline model with the same architecture (the far left orange point) by a large margin on all metrics while requiring far fewer FLOPs. This is due to the additional evidence per training example that we get when training models with distillation losses. Notably, the minADE and miss rate metrics of the distilled model are better than the teacher ensemble. The distillation result for the Argoverse dataset is shown by the green dots and lines in fig \ref{fig:argoverse}. Here we see that the distilled models also perform far better than the base model. For the Argoverse experiments we trained one model with 0.75X of the FLOPs required by the base model and one with required FLOPs equal to the base model. Both models show significant performance improvement over the baseline model (the far left orange dot). Despite having 25\% fewer required FLOPs, the smaller model still  significantly outperforms the baseline model. The key result is that training a distilled model from a much larger ensemble provides a significant increase in performance with much lower compute cost than the ensemble, remaining suitable for running onboard the robot.

\begin{figure}[h]
\centering
\label{fig:qualitative}
    \includegraphics[width=\linewidth]{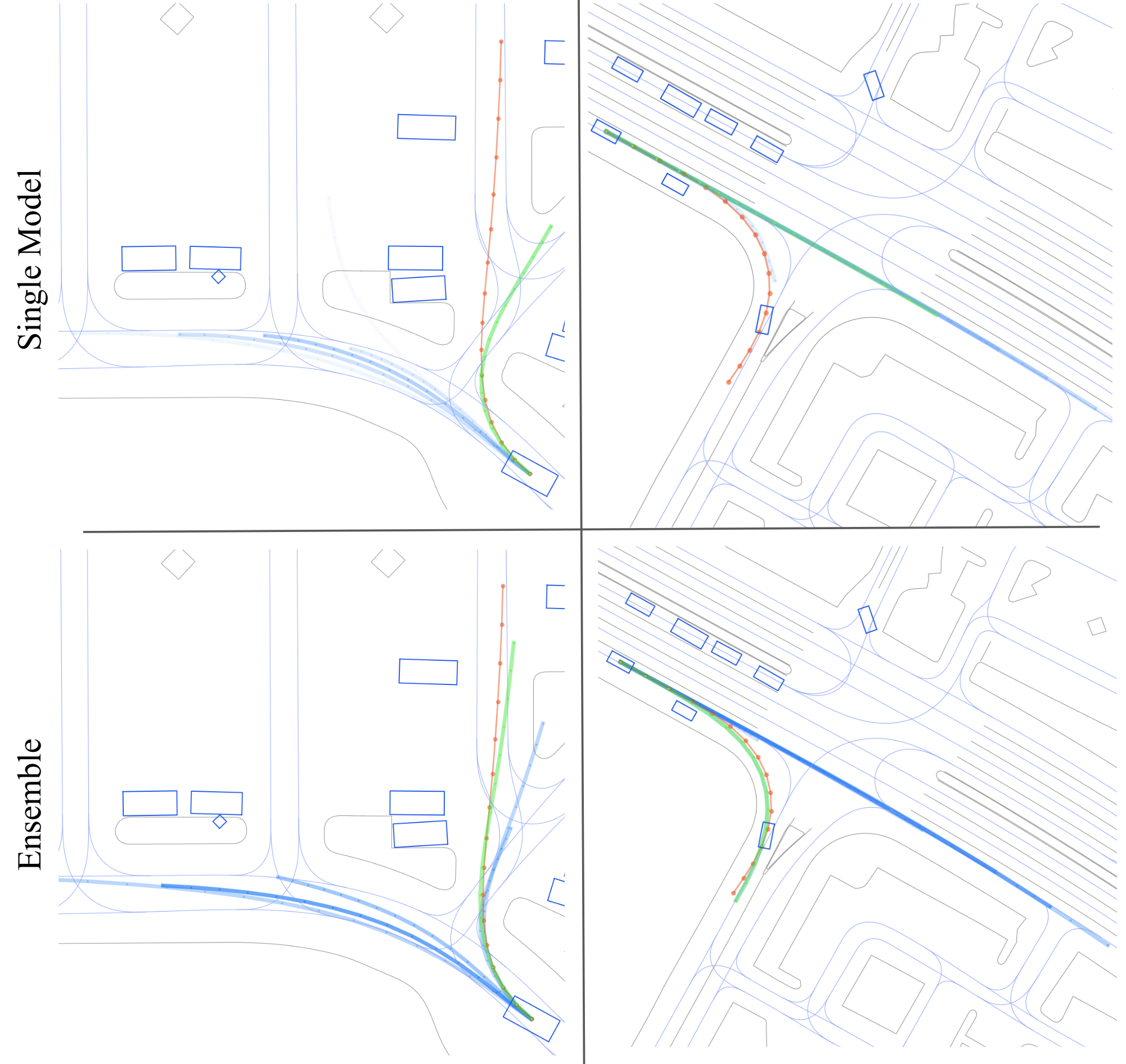}
\label{fig:qualitative}    
\caption{2 Qualitative examples for a single teacher models (top row) compared with a $K=28$ teacher ensemble (bottom row). Both setups output 6 trajectories after NMS. The red trajectory is the groundtruth, the green is the prediction closest to the groundtruth, and the blue are the remaining predictions. The grid lines represents the background scene including the roadgraph and other agents.} \label{fig:qualitative}
\end{figure}

Table \ref{tab:sota} shows a comparison of our models with the current WOMD leaderboard. Our ensemble model ranks third on the leaderboard. The distilled model has reduced performance on most metrics but with over 20X fewer FLOPs required. It is not known at this time how the other entries on the leaderboard compare to the distilled model in terms of required FLOPs. Notably the distilled model has the best minADE performance of any model on the leaderboard. The key result is that the distilled model provides high performance in a small model that can be deployed in real-time onboard robotics applications.

\section {Conclusion}

We present a novel framework for ensemble distillation for a multi-modal output distribution domain of trajectory prediction. Our WOMD ensemble ranks second on the leaderboard while our Argoverse ensemble ranks 3rd. Our distilled student from the WOMD ensemble model provides high performance while requiring over 20X fewer FLOPS to compute than the ensemble model. We show that the same trends hold on the Argoverse dataset for both ensembles and distilled models. The Argoverse ensemble model ranks third on the leaderboard and the distilled model significantly improves over the baseline. These experiments demonstrate ensemble distillation as an effective method for improving performance for robotics applications with limited onboard compute budgets.

{\small
\bibliographystyle{ieee_fullname}
\bibliography{egbib}

\begin{thebibliography}{10}\itemsep=-1pt

\bibitem{social-lstm}
Alexandre Alahi, Kratarth Goel, Vignesh Ramanathan, Alexandre Robicquet, Li
  Fei-Fei, and Silvio Savarese.
\newblock Social lstm: Human trajectory prediction in crowded spaces.
\newblock {\em 2016 IEEE Conference on Computer Vision and Pattern Recognition
  (CVPR)}, pages 961--971, 2016.

\bibitem{Allen-Zhu2020Distill}
Zeyuan Allen-Zhu and Yuanzhi Li.
\newblock Towards understanding ensemble, knowledge distillation and
  self-distillation in deep learning, 2020.

\bibitem{allen-zhu2020towards}
Zeyuan Allen-Zhu, Yuanzhi Li, and Zeyuan Allen-Zhu.
\newblock Towards understanding ensemble, knowledge distillation and
  self-distillation in deep learning.
\newblock {\em arXiv preprint}, (2012.09816), December 2020.

\bibitem{Breiman2004StackedR}
L. Breiman.
\newblock Stacked regressions.
\newblock {\em Machine Learning}, 24:49--64, 2004.

\bibitem{casas2020spagnn}
Sergio Casas, Cole Gulino, Renjie Liao, and Raquel Urtasun.
\newblock Spagnn: Spatially-aware graph neural networks for relational behavior
  forecasting from sensor data.
\newblock In {\em IEEE Intl. Conf. on Robotics and Automation}. IEEE, 2020.

\bibitem{DBLP:journals/corr/abs-2101-07907}
Sergio Casas, Wenjie Luo, and Raquel Urtasun.
\newblock Intentnet: Learning to predict intention from raw sensor data.
\newblock {\em CoRR}, abs/2101.07907, 2021.

\bibitem{multipath}
Yuning Chai, Benjamin Sapp, Mayank Bansal, and Dragomir Anguelov.
\newblock Multipath: Multiple probabilistic anchor trajectory hypotheses for
  behavior prediction.
\newblock In {\em CoRL}, 2019.

\bibitem{argoverse}
Ming{-}Fang Chang, John Lambert, Patsorn Sangkloy, Jagjeet Singh, Slawomir Bak,
  Andrew Hartnett, De Wang, Peter Carr, Simon Lucey, Deva Ramanan, and James
  Hays.
\newblock Argoverse: 3d tracking and forecasting with rich maps.
\newblock {\em CoRR}, abs/1911.02620, 2019.

\bibitem{cui2019mtp}
Henggang Cui, Vladan Radosavljevic, Fang-Chieh Chou, Tsung-Han Lin, Thi Nguyen,
  Tzu-Kuo Huang, Jeff Schneider, and Nemanja Djuric.
\newblock Multimodal trajectory predictions for autonomous driving using deep
  convolutional networks.
\newblock In {\em 2019 International Conference on Robotics and Automation
  (ICRA)}, pages 2090--2096. IEEE, 2019.

\bibitem{Dietterich2000EnsembleMI}
Thomas~G. Dietterich.
\newblock Ensemble methods in machine learning.
\newblock In {\em Multiple Classifier Systems}, 2000.

\bibitem{WOMD}
Scott Ettinger, Shuyang Cheng, Benjamin Caine, Chenxi Liu, Hang Zhao, Sabeek
  Pradhan, Yuning Chai, Benjamin Sapp, Charles~R. Qi, Yin Zhou, Zoey Yang,
  Aurelien Chouard, Pei Sun, Jiquan Ngiam, Vijay Vasudevan, Alexander McCauley,
  Jonathon Shlens, and Dragomir Anguelov.
\newblock Large scale interactive motion forecasting for autonomous driving :
  The waymo open motion dataset.
\newblock {\em CoRR}, abs/2104.10133, 2021.

\bibitem{Ettinger2021WOMD}
Scott Ettinger, Shuyang Cheng, Benjamin Caine, Chenxi Liu, Hang Zhao, Sabeek
  Pradhan, Yuning Chai, Benjamin Sapp, Charles~R. Qi, Yin Zhou, Zoey Yang,
  Aurelien Chouard, Pei Sun, Jiquan Ngiam, Vijay Vasudevan, Alexander McCauley,
  Jonathon Shlens, and Dragomir Anguelov.
\newblock Large scale interactive motion forecasting for autonomous driving :
  The waymo open motion dataset.
\newblock {\em CoRR}, abs/2104.10133, 2021.

\bibitem{Fort2019DeepEA}
Stanislav Fort, Huiyi Hu, and Balaji Lakshminarayanan.
\newblock Deep ensembles: A loss landscape perspective.
\newblock {\em ArXiv}, abs/1912.02757, 2019.

\bibitem{Furlanello2018BornAN}
Tommaso Furlanello, Zachary~Chase Lipton, Michael Tschannen, Laurent Itti, and
  Anima Anandkumar.
\newblock Born again neural networks.
\newblock In {\em ICML}, 2018.

\bibitem{gao2020vectornet}
Jiyang Gao, Chen Sun, Hang Zhao, Yi Shen, Dragomir Anguelov, Congcong Li, and
  Cordelia Schmid.
\newblock {VectorNet}: Encoding hd maps and agent dynamics from vectorized
  representation.
\newblock In {\em CVPR}, 2020.

\bibitem{gu21dense_tnt}
Junru Gu, Chen Sun, and Hang Zhao.
\newblock Densetnt: End-to-end trajectory prediction from dense goal sets.
\newblock {\em CoRR}, abs/2108.09640, 2021.

\bibitem{social-gan}
Agrim Gupta, Justin Johnson, Li Fei-Fei, Silvio Savarese, and Alexandre Alahi.
\newblock Social gan: Socially acceptable trajectories with generative
  adversarial networks.
\newblock In {\em Proceedings of the IEEE Conference on Computer Vision and
  Pattern Recognition (CVPR)}, June 2018.

\bibitem{Hansen1990NeuralNE}
Lars~Kai Hansen and Peter Salamon.
\newblock Neural network ensembles.
\newblock {\em IEEE Trans. Pattern Anal. Mach. Intell.}, 12:993--1001, 1990.

\bibitem{Hinton2015DistillingTK}
Geoffrey~E. Hinton, Oriol Vinyals, and Jeffrey Dean.
\newblock Distilling the knowledge in a neural network.
\newblock {\em ArXiv}, abs/1503.02531, 2015.

\bibitem{DBLP:journals/corr/abs-1906-08945}
Joey Hong, Benjamin Sapp, and James Philbin.
\newblock Rules of the road: Predicting driving behavior with a convolutional
  model of semantic interactions.
\newblock {\em CoRR}, abs/1906.08945, 2019.

\bibitem{lyft5}
John Houston, Guido Zuidhof, Luca Bergamini, Yawei Ye, Ashesh Jain, Sammy
  Omari, Vladimir Iglovikov, and Peter Ondruska.
\newblock One thousand and one hours: Self-driving motion prediction dataset.
\newblock {\em CoRR}, abs/2006.14480, 2020.

\bibitem{scalinglaws1}
Jared Kaplan, Sam McCandlish, T.~J. Henighan, Tom~B. Brown, Benjamin Chess,
  Rewon Child, Scott Gray, Alec Radford, Jeff Wu, and Dario Amodei.
\newblock Scaling laws for neural language models.
\newblock {\em ArXiv}, abs/2001.08361, 2020.

\bibitem{wimp2020}
Siddhesh Khandelwal, William Qi, Jagjeet Singh, Andrew Hartnett, and Deva
  Ramanan.
\newblock What-if motion prediction for autonomous driving.
\newblock {\em ArXiv}, 2020.

\bibitem{Kim2016SequenceLevelKD}
Yoon Kim and Alexander~M. Rush.
\newblock Sequence-level knowledge distillation.
\newblock In {\em EMNLP}, 2016.

\bibitem{Krogh1994NeuralNE}
Anders Krogh and Jesper Vedelsby.
\newblock Neural network ensembles, cross validation, and active learning.
\newblock In {\em NIPS}, 1994.

\bibitem{DBLP:journals/corr/LeeCVCTC17}
Namhoon Lee, Wongun Choi, Paul Vernaza, Christopher~B. Choy, Philip H.~S. Torr,
  and Manmohan~Krishna Chandraker.
\newblock {DESIRE:} distant future prediction in dynamic scenes with
  interacting agents.
\newblock {\em CoRR}, abs/1704.04394, 2017.

\bibitem{Li2014LearningSD}
Jinyu Li, Rui Zhao, Jui~Ting Huang, and Yifan Gong.
\newblock Learning small-size dnn with output-distribution-based criteria.
\newblock In {\em INTERSPEECH}, 2014.

\bibitem{liang2020laneGCN}
Ming Liang, Bin Yang, Rui Hu, Yun Chen, Renjie Liao, Song Feng, and Raquel
  Urtasun.
\newblock Learning lane graph representations for motion forecasting.
\newblock {\em arXiv preprint arXiv:2007.13732}, 2020.

\bibitem{adamw}
Ilya Loshchilov and Frank Hutter.
\newblock Decoupled weight decay regularization.
\newblock In {\em ICLR}, 2019.

\bibitem{Maclin1999PopularEM}
R. Maclin and David~W. Opitz.
\newblock Popular ensemble methods: An empirical study.
\newblock {\em J. Artif. Intell. Res.}, 11:169--198, 1999.

\bibitem{Malinin2019Ensemble}
Andrey Malinin, Bruno Mlodozeniec, and Mark Gales.
\newblock Ensemble distribution distillation, 2019.

\bibitem{multiheadAF}
Jean~Pierre Mercat, Thomas Gilles, Nicole~El Zoghby, Guillaume Sandou,
  Dominique Beauvois, and Guillermo~Pita Gil.
\newblock Multi-head attention for multi-modal joint vehicle motion
  forecasting.
\newblock {\em 2020 IEEE International Conference on Robotics and Automation
  (ICRA)}, pages 9638--9644, 2020.

\bibitem{Nayakanti2022Wayformer}
Nigamaa Nayakanti, Rami Al-Rfou, Aurick Zhou, Kratarth Goel, Khaled~S. Refaat,
  and Benjamin Sapp.
\newblock Wayformer: Motion forecasting via simple \& efficient attention
  networks.
\newblock {\em ArXiv}, abs/2207.05844, 2022.

\bibitem{rhinehart2019precog}
Nicholas Rhinehart, Rowan McAllister, Kris Kitani, and Sergey Levine.
\newblock Precog: Prediction conditioned on goals in visual multi-agent
  settings.
\newblock In {\em ECCV}, 2019.

\bibitem{trajectronpp}
Tim Salzmann, Boris Ivanovic, Punarjay Chakravarty, and Marco Pavone.
\newblock Trajectron++: Multi-agent generative trajectory forecasting with
  heterogeneous data for control.
\newblock {\em CoRR}, abs/2001.03093, 2020.

\bibitem{sapp2019multipath}
Benjamin Sapp, Yuning Chai, Mayank Bansal, and Dragomir Anguelov.
\newblock {MultiPATH}: Multiple probabilistic anchor trajectory hypotheses for
  behavior prediction.
\newblock In {\em Conf. on Robot Learning}, 2019.

\bibitem{Schapire1997BoostingTM}
Robert~E. Schapire, Yoav Freund, Peter Barlett, and Wee~Sun Lee.
\newblock Boosting the margin: A new explanation for the effectiveness of
  voting methods.
\newblock In {\em ICML}, 1997.

\bibitem{shi2022mtr}
Shaoshuai Shi, Li Jiang, Dengxin Dai, and Bernt Schiele.
\newblock Motion transformer with global intention localization and local
  movement refinement.
\newblock {\em Advances in Neural Information Processing Systems}, 2022.

\bibitem{Su2022NarrowingTC}
DiJia Su, Bertrand Douillard, Rami Al-Rfou, Cheol~Soo Park, and Benjamin Sapp.
\newblock Narrowing the coordinate-frame gap in behavior prediction models:
  Distillation for efficient and accurate scene-centric motion forecasting.
\newblock {\em 2022 International Conference on Robotics and Automation
  (ICRA)}, pages 653--659, 2022.

\bibitem{multipathpp}
Balakrishnan Varadarajan, Ahmed~S. Hefny, Avikalp Srivastava, Khaled~S. Refaat,
  Nigamaa Nayakanti, Andre Cornman, Kan Chen, Bertrand Douillard, C.~P. Lam,
  Drago Anguelov, and Benjamin Sapp.
\newblock Multipath++: Efficient information fusion and trajectory aggregation
  for behavior prediction.
\newblock In {\em ICRA}, 2021.

\bibitem{vaswani2017attention}
Ashish Vaswani, Noam Shazeer, Niki Parmar, Jakob Uszkoreit, Llion Jones,
  Aidan~N Gomez, {\L}ukasz Kaiser, and Illia Polosukhin.
\newblock Attention is all you need.
\newblock In {\em NeurIPS}, 2017.

\bibitem{wilson2023argoverse}
Benjamin Wilson, William Qi, Tanmay Agarwal, John Lambert, Jagjeet Singh,
  Siddhesh Khandelwal, Bowen Pan, Ratnesh Kumar, Andrew Hartnett,
  Jhony~Kaesemodel Pontes, et~al.
\newblock Argoverse 2: Next generation datasets for self-driving perception and
  forecasting.
\newblock {\em arXiv preprint arXiv:2301.00493}, 2023.

\bibitem{zhao2020tnt}
Hang Zhao, Jiyang Gao, Tian Lan, Chen Sun, Benjamin Sapp, Balakrishnan
  Varadarajan, Yue Shen, Yi Shen, Yuning Chai, Cordelia Schmid, et~al.
\newblock Tnt: Target-driven trajectory prediction.
\newblock {\em arXiv preprint arXiv:2008.08294}, 2020.

\bibitem{Zhou2020UnderstandingKD}
Chunting Zhou, Graham Neubig, and Jiatao Gu.
\newblock Understanding knowledge distillation in non-autoregressive machine
  translation.
\newblock {\em ArXiv}, abs/1911.02727, 2020.

\end{thebibliography}
}

\clearpage
\newpage
\pagebreak
\newpage
\appendix

\subsection{Distillation temperature analysis}

\begin{figure}[h]
  \centering
  \includegraphics[width=0.8\linewidth]{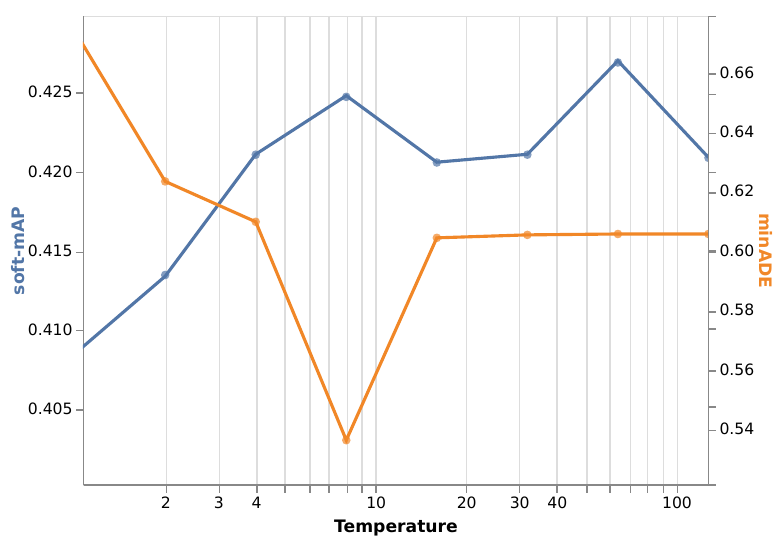}
  \caption{The temperature $(\tau)$ of distillation is an important parameter to balance the tradeoff between soft-mAP and minADE. Results show distillation of a 28 model ensemble.}
  \label{fig:temperature}
\end{figure}

A challenge in training the student is to retain diversity while still performing well on the common case. Increasing the temperature parameter promotes diversity of learned trajectories by flattening the distribution. Figure \ref{fig:temperature} shows the results of a study varying the distillation temperature parameter in models distilled from the $K=28$ ensemble using the losses described in section \ref{section:losses} with the simplification described in \ref{section:approximation} and $w_{gt}=0.4$, $w_{ent}=1.0$, and $w_{var}=0.1$.  For these experiments, soft-mAP (higher is better) is a proxy for diversity while minADE (lower is better) is a proxy for common case performance. The experiments show an optimum at $\tau=8$. 

\subsection{Distillation Loss Analysis}
We perform analysis of different formulations of our distillation loss. Here we ablate the use of the bijective mapping decribed in section \ref{section:special} denoted by $\pi_{1:1}$. The teacher models always output $N_T = 64$ trajectories prior to NMS. In this analysis use an ensemble of $K=28$ teacher models. Thus we get a total of $K*N_T$ trajectories pre-NMS. These are the NMS-ed down to $M_T$ trajectories. The student models output either $N_S=6$ trajectories or $K=64$ trajectories, which are then NMS-ed down to $M_S=6$ (when $N_S > M_S$) trajectories for metric computation. When $M_T = N_T$, we also use the bijective mapping and compare the results with not using it. 

\begin{table}[h]
    \centering
    \resizebox{\linewidth}{!}{
        \begin{tabular}{cc|c|cc}
        \toprule
        $M_T$ & $N_S$ & $\pi_{1:1}$ &  minADE  ($\downarrow$) & mAP ($\uparrow$) \\ 
        \midrule
        6 & 6 & \checkmark & 0.64 & 0.37 \\
        64 & 64 & \checkmark & 0.55 & \textbf{0.43} \\
        64 & 6 & & \textbf{0.53} & \textbf{0.43} \\
        \bottomrule
        \end{tabular}
    }
    \caption{Comparison of the different distillation formulations in terms of trajectories fed into the model, output from the model and the distillation loss used for distillation on WOMD validation set. The number of teacher models $K=28$ to form the ensemble, the number of output modes from each teacher model is $N_T=64$ and the number of output trajectories from each student post NMS $M_S=6$ for metric computation.}
    \label{tab:compare_loss}
\end{table}

\begin{figure}[h]
\centering
\label{fig:results}
    \includegraphics[trim={0 0 0 2cm}, width=0.49\linewidth]{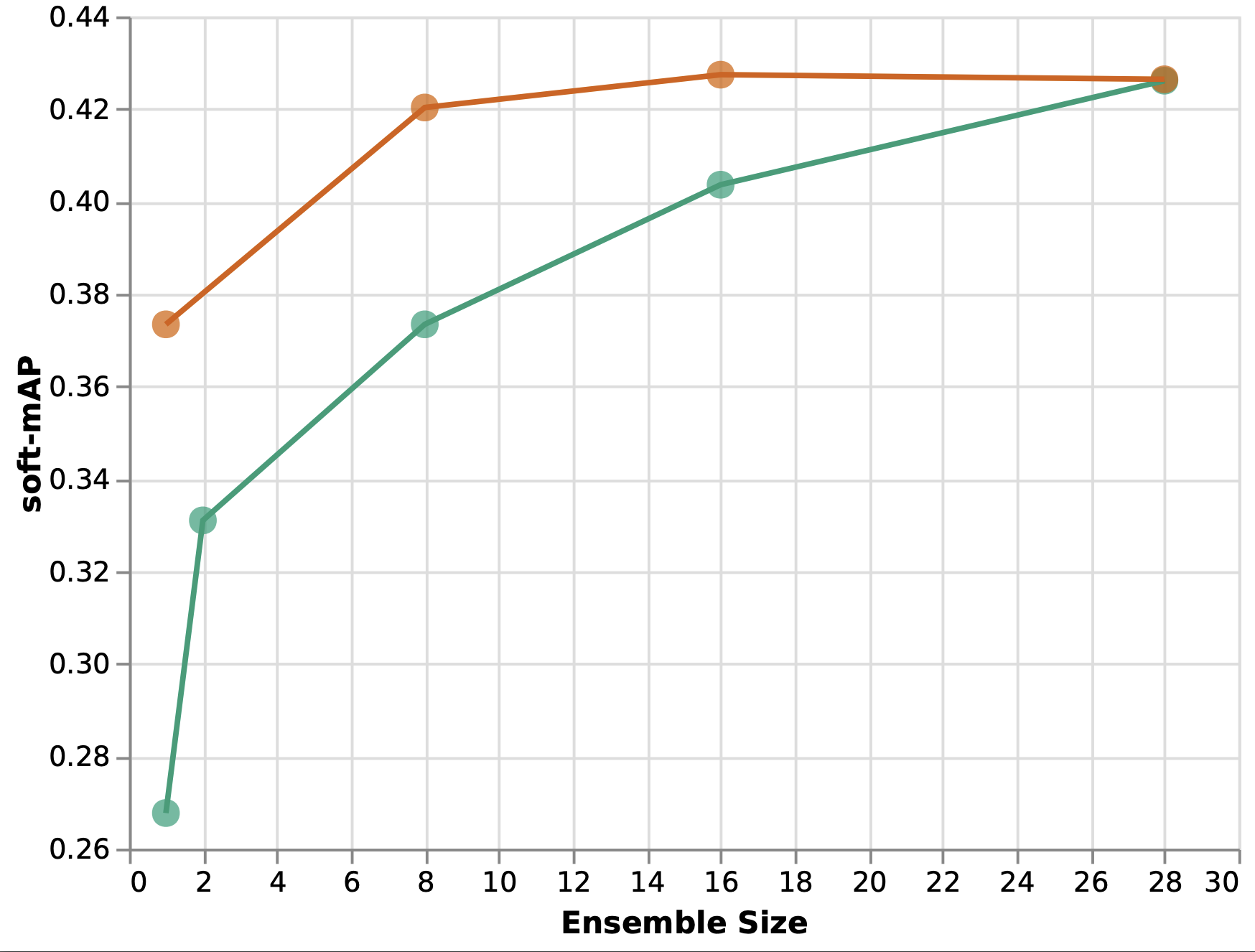}
    \includegraphics[trim={0 0 0 2cm},width=0.49\linewidth]{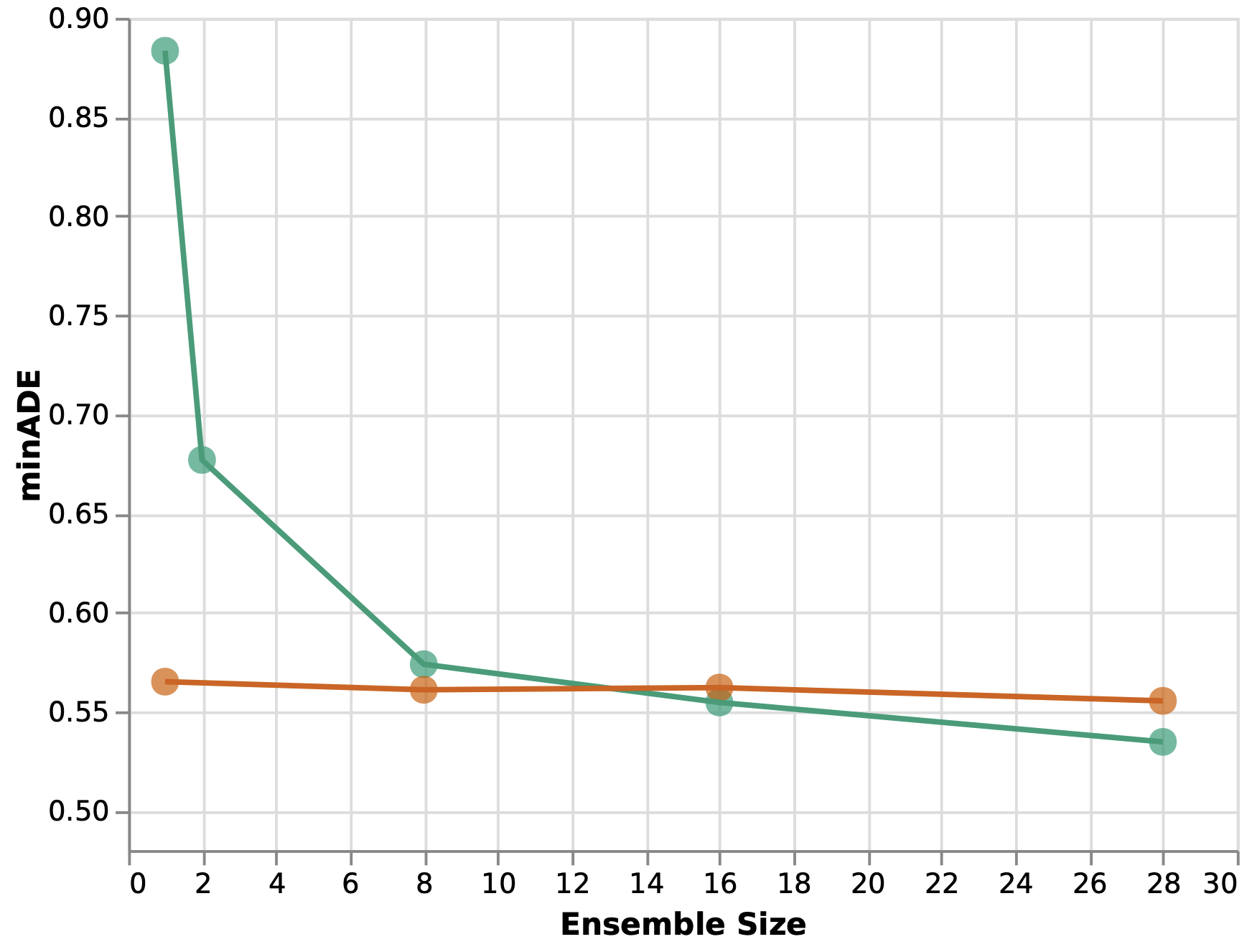}
\caption{ soft-mAP and MinADE metric results for the using students distilled with the bijective mapping $\pi_{1:1}$ (orange) and without (green) as a function of the number of teacher models $K$ in the teacher ensemble. } \label{fig:compare_for_xent_vs_gmm}
\end{figure}

 The results are tabulated in table \ref{tab:compare_loss}. We see it is better for distillation models to have teachers that are more expressive and output more modes in their distribution. Thus, $M_T = 64$ models output models with $M_T=6$. It is also interesting to note that while the overall best results are obtained by the model (row 3) that is allowed to learn the mapping between the modes of the student $N_S$ and the modes of the teachers $M_T$, the bijective mapping proposed in section \ref{section:special} does relatively reasonably and it not too far off in terms of the overall best performance (row 2).
 
 Figures \ref{fig:compare_for_xent_vs_gmm} show the results of experiments for two student models with the same architecture $N_S=64$, $M_S=64$ with teachers trained with $N_T=64$ in an ensemble. The number of teachers or ensemble size $K$ is varied from 1 ( self-distillation) to 28. The output number of modes from the ensemble is also NMS-ed down to $M_T=64$. We then either use the loss with $\pi_{1:1}$ (orange) or without (green) and see the effect of scaling up the ensemble size. We observe that for both of the loss formulations adding more models to the ensemble from which models are distilled, cause the performance to improve. In addition we see that using the bijective mapping  between teacher modes and student modes is better for small ensembles, but with a large enough ensemble size allowing the model to learn the mapping is better. This ultimately results in our state of the art model, although the deterministic bijective mapping is very competitive. 

\subsection{Qualitative Analysis}
In this section we presentimages of qualitative differences between the single SoTA model and the Ensemble model with $K = 28$ teachers. For all following figures Fig \ref{fig:first_qual} - \ref{fig:last_qual} both the ensemble and the single model output 6 trajectories after NMS. For all figures the red trajectory is the groundtruth, the green is the prediction closest to the groundtruth, and the blue are the remaining predictions with higher opacity representing higher confidence. While the ensemble may not improve on every example, it does perform better on average as shown in the quantitative metrics in section \ref{sec:experiments}.  Here we present a few examples where the ensemble model performs better than the single model.

\begin{figure*}[b]
\centering
    \includegraphics[width=\linewidth]{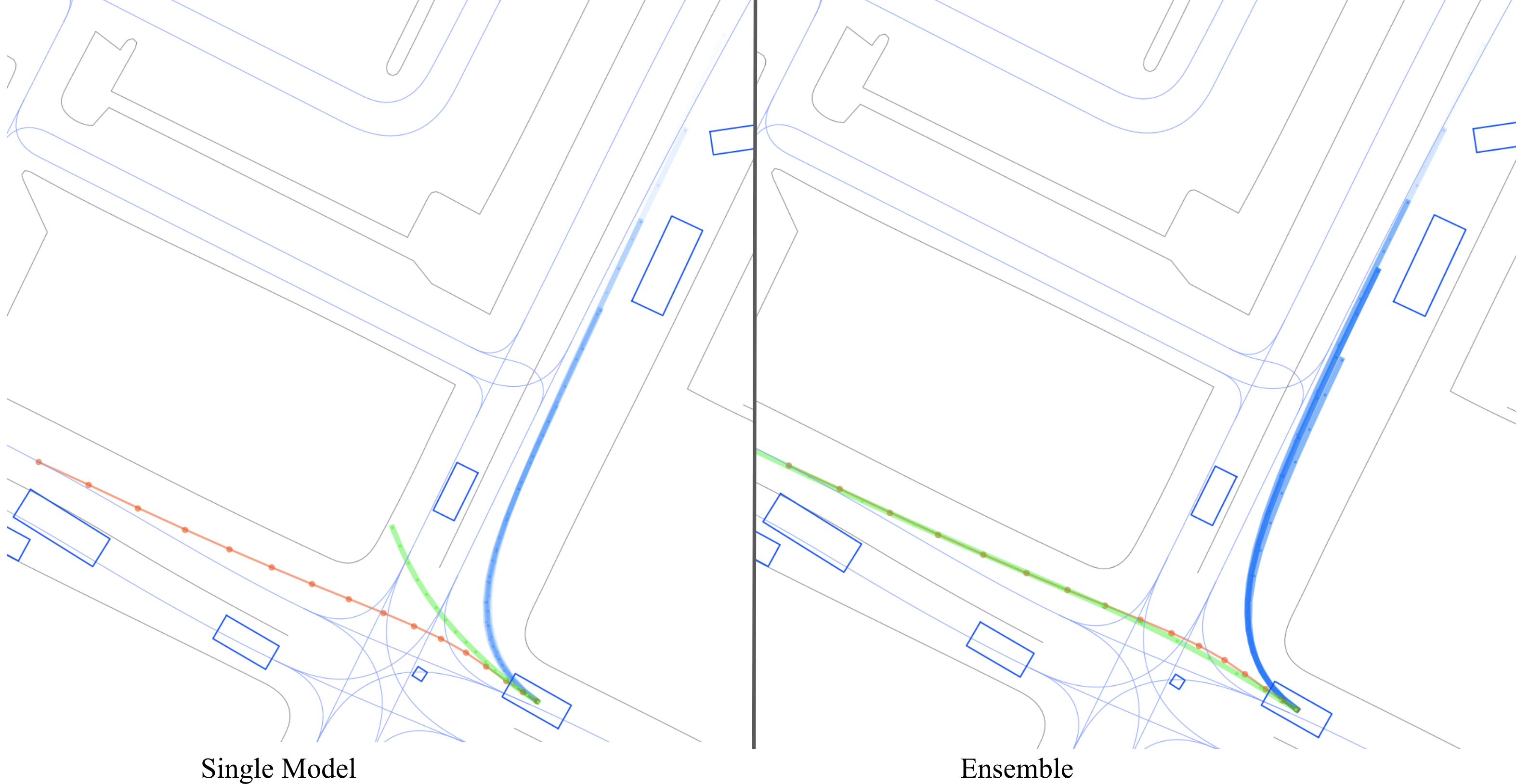}
\caption{Here the ensemble correctly predicts a forward trajectory while eliminating the unlikely path that drives off the road predicted by the single model.} 
\label{fig:first_qual}
\end{figure*}

\begin{figure*}[b]
\centering
    \includegraphics[width=\linewidth]{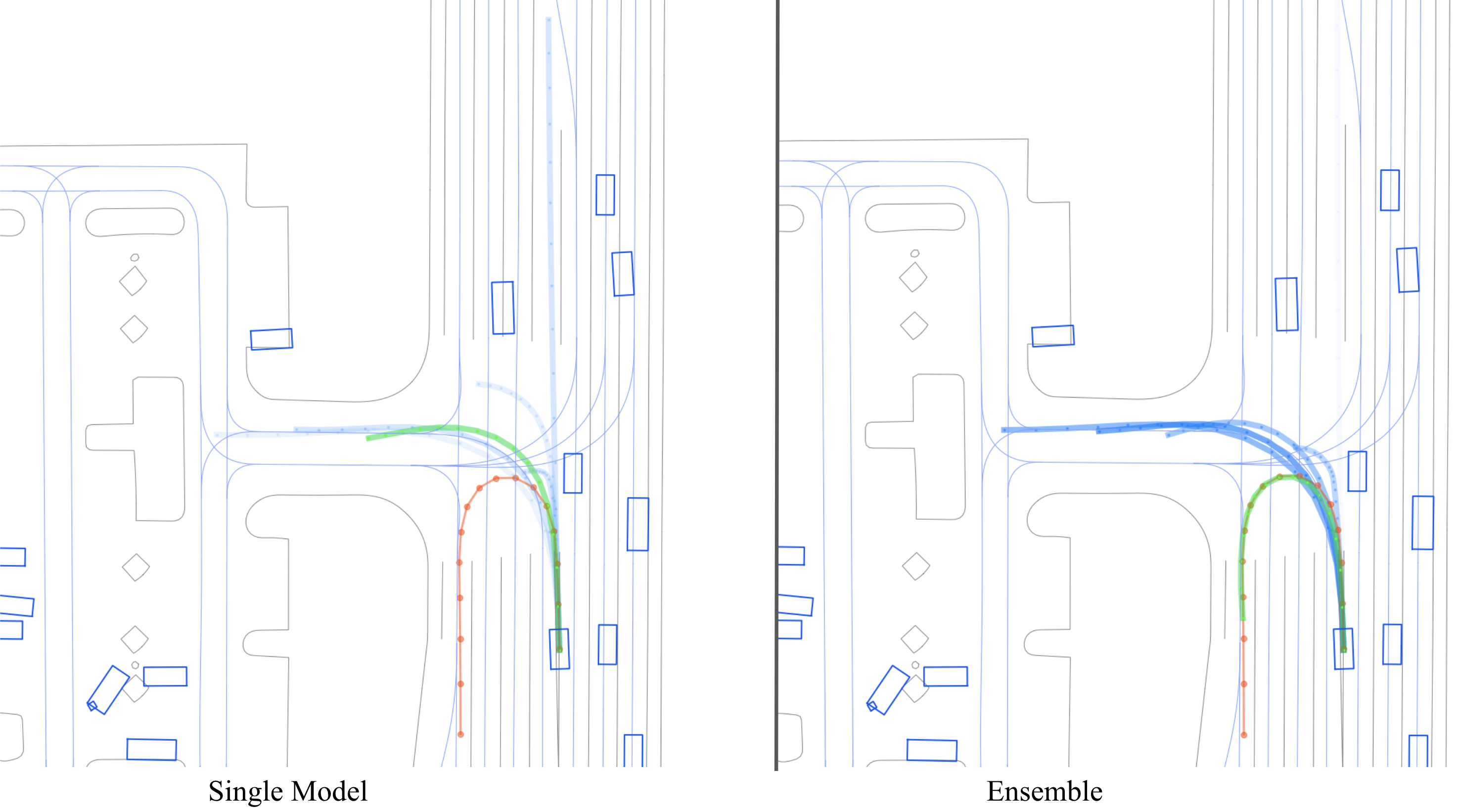}
\caption{Here we see that the ensemble correctly predicts that making a U-turn or a left turn is more likely from this lane as there is no lane continuing forward from the vehicle's position.}
\end{figure*}

\begin{figure*}[b]
\centering
    \includegraphics[width=\linewidth]{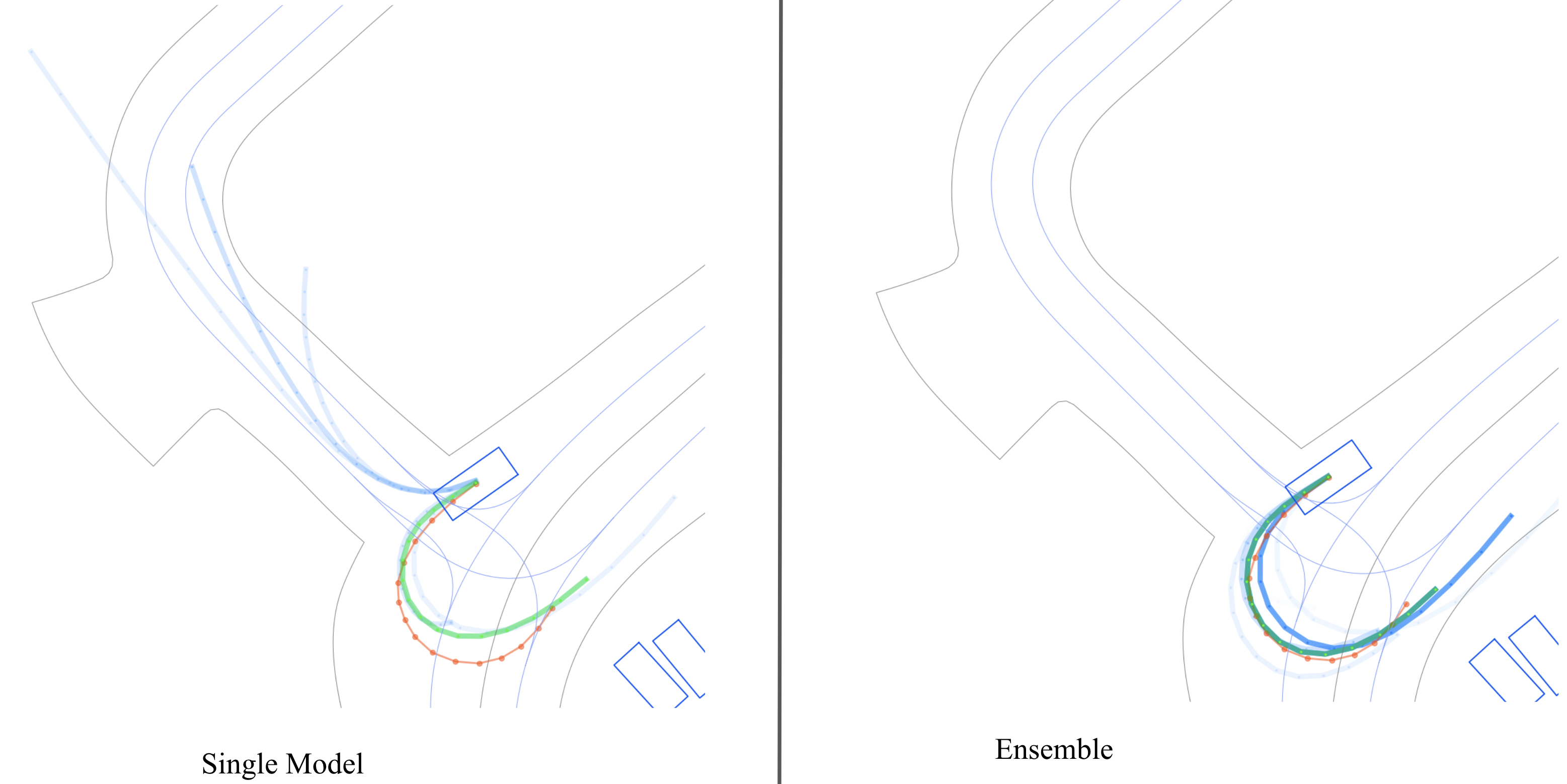}
\caption{In this example, the ensemble confidently predicts a U-turn based on the past trajectory of this vehicle.} 
\end{figure*}

\begin{figure*}[b]
\centering
    \includegraphics[width=\linewidth]{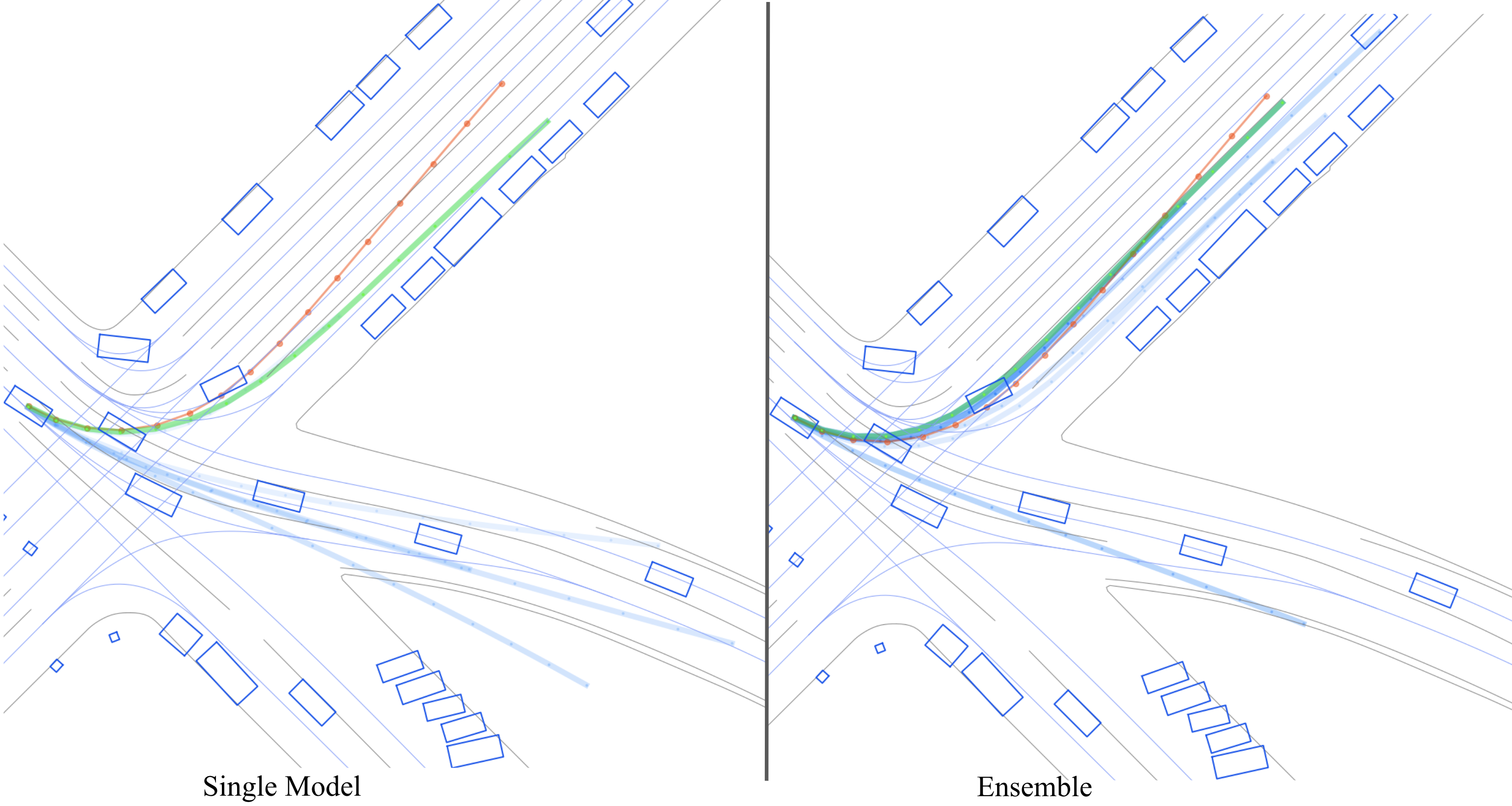}
\caption{In this case, the ensemble has more confidence in the correct output mode based on the past history of the vehicle.} 
\end{figure*}

\begin{figure*}[b]
\centering
    \includegraphics[width=\linewidth]{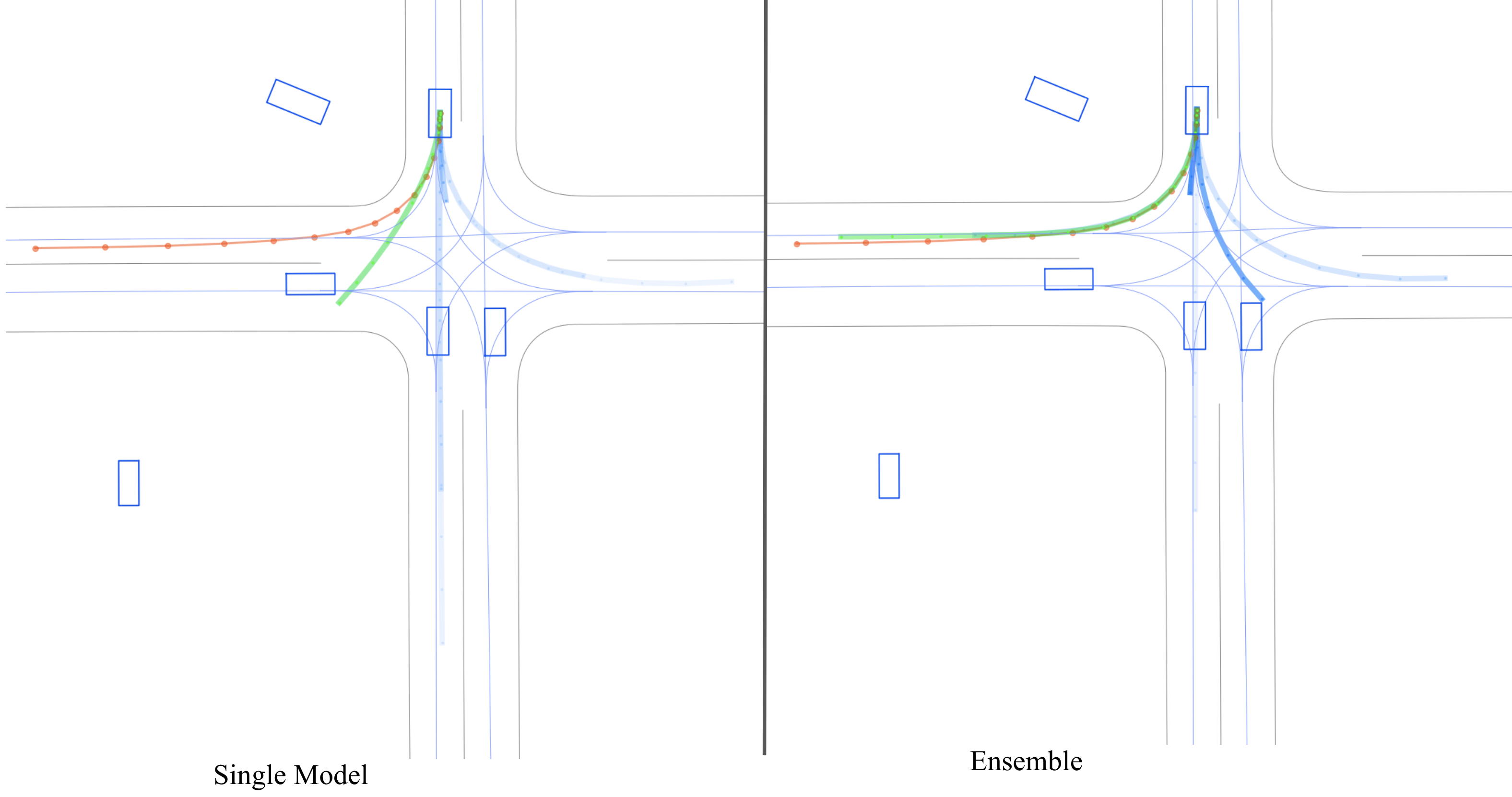}
\caption{Here the ensemble predicts a trajectory that better matches the ground truth mode while the single model has no prediction to the right lane.} 
\end{figure*}

\begin{figure*}[b]
\centering
    \includegraphics[width=\linewidth]{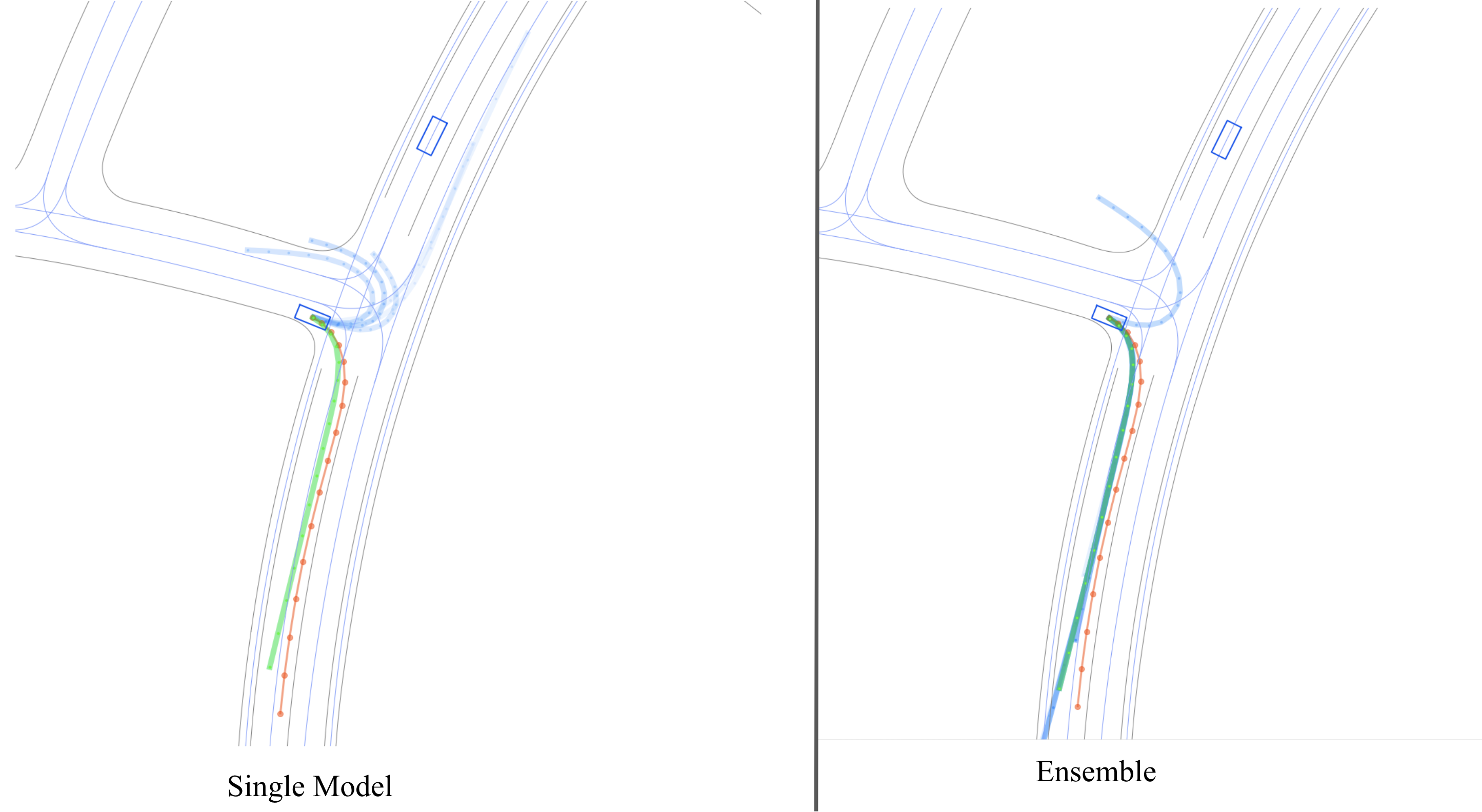}
\caption{Here the ensemble correctly predicts that a right turn is much more likely than a u-turn or a left turn from this position.} 
\label{fig:last_qual}
\end{figure*}

\end{document}